\documentclass[sigconf]{acmart}
\AtBeginDocument{%
  }

\usepackage{balance}
\usepackage{booktabs}
\usepackage{makecell}
\usepackage{enumitem}
\usepackage{multirow}
\usepackage{array}
\usepackage{arydshln}
\usepackage{nicematrix}

\copyrightyear{2026}
\acmYear{2026}
\setcopyright{cc}
\setcctype{by}
\acmConference[KDD 2026] {Proceedings of the 32nd ACM SIGKDD Conference on Knowledge Discovery and Data Mining V.2}{August 9--13, 2026}{Jeju Island, Republic of Korea.}
\acmBooktitle{Proceedings of the 32nd ACM SIGKDD Conference on Knowledge Discovery and Data Mining V.2 (KDD 2026), August 9--13, 2026, Jeju Island, Republic of Korea}
\acmISBN{979-8-4007-2259-2/2026/08}
\acmDOI{10.1145/3770855.3817734}

\makeatletter
\def\@fnsymbol#1{%
  \ensuremath{%
    \ifcase#1\or
      \dagger\or
      *\or
      \ddagger\or
      \mathsection\or
      \mathparagraph\or
      \|\or
      **\or
      \dagger\dagger\or
      \ddagger\ddagger
    \else
      \@ctrerr
    \fi
  }%
}
\makeatother

\begin{document}
  
\title[$\text{GPH}^{2}$]{Unified Multi-Domain Graph Pre-training for Homogeneous and Heterogeneous Graphs via Domain-Specific Expert Encoding}

\author{Chundong Liang}
\authornote{Both authors contributed equally to this research.}
\email{liangchundong@tju.edu.cn}
\affiliation{%
  \department{School of Computer Science and Technology}
  \institution{Tianjin University}
  \city{Tianjin}
  \country{China}}
\affiliation{%
  \institution{North China University of Science and Technology}
  \city{Tangshan}
  \country{China}}

\author{Yongqi Huang}
\authornotemark[1]
\email{yqhuang@tju.edu.cn}
\affiliation{%
  \department{School of Computer Science and Technology}
  \institution{Tianjin University}
  \city{Tianjin}
  \country{China}}

\author{Dongxiao He}
\authornote{Corresponding author.}
\author{Peiyuan Li}
\email{hedongxiao@tju.edu.cn}
\email{lipeiyuan04@tju.edu.cn}
\affiliation{%
  \department{School of Computer Science and Technology}
  \institution{Tianjin University}
  \city{Tianjin}
  \country{China}}

\author{Yawen Li}
\email{warmly0716@126.com}
\affiliation{%
  \department{School of Economics and Management}
  \institution{Beijing University of Posts and Telecommunications}
  \city{Beijing}
  \country{China}}

\author{Di Jin}
\email{jindi@tju.edu.cn}
\affiliation{%
  \department{School of Computer Science and Technology}
  \institution{Tianjin University}
  \city{Tianjin}
  \country{China}}

\author{Weixiong Zhang}
\email{weixiong.zhang@polyu.edu.hk}
\affiliation{%
  \department{Departments of Health Technology \& Informatics; Data Science \& Artificial Intelligence; and Computing}
  \institution{\mbox{The Hong Kong Polytechnic University}}
  \city{Kowloon}
  \country{Hong Kong}}

\renewcommand{\shortauthors}{Chundong Liang, Yongqi Huang et al.}

\begin{abstract}
Graph pre-training has achieved remarkable success in recent years, delivering transferable representations for downstream adaptation. However, most existing methods are designed for either homogeneous or heterogeneous graphs, thereby hindering unified graph modeling across diverse graph types. This separation contradicts real-world applications, where mixed homogeneous and heterogeneous graphs are ubiquitous, and distribution shifts between upstream pre-training and downstream deployment are common. In this paper, we empirically demonstrate that a balanced mixture of homogeneous and heterogeneous graph pre-training benefits downstream tasks and propose a unified multi-domain \textbf{G}raph \textbf{P}re-training method across \textbf{H}omogeneous and \textbf{H}eterogeneous graphs ($\mathbf{GPH^{2}}$). To address the lack of a unified encoder for homogeneous and heterogeneous graphs, we propose a Unified Multi-View Graph Construction that simultaneously encodes both without explicit graph-type-specific designs. To cope with the increased cross-domain distribution discrepancies arising from mixed graphs, we introduce domain-specific expert encoding. Each expert is independently pre-trained on a single graph to capture domain-specific knowledge, thereby shielding the pre-training encoder from the adverse effects of cross-domain discrepancies. For downstream tasks, we further design a Task-oriented Expert Fusion Strategy that adaptively integrates multiple experts based on their discriminative strengths. Extensive experiments on mixed graphs demonstrate that $\text{GPH}^{2}$ enables stable transfer across graph types and domains, significantly outperforming existing graph pre-training methods.
\end{abstract}




\maketitle

\section{Introduction}\label{sec:intro}

Graphs are ubiquitous non-Euclidean data structures widely used to model complex real-world systems, such as social networks~\cite{Social_nets}, citation networks~\cite{Cora}, and molecular graphs~\cite{KANO}. Graph Representation Learning (GRL) aims to learn representations for graph data, enabling effective downstream applications~\cite{GNNsurvey1}. While early studies mainly relied on supervised training, their performance often depends on abundant labeled data and may generalize poorly across tasks and domains. Inspired by advances in self-supervised learning~\cite{SSLsurvey1, SSLsurvey2, SGRL}, the "pre-train and then fine-tune" paradigm has become increasingly prevalent: an encoder is first pre-trained on a large number of unlabeled graphs to learn general graph representations and then adapted to specific downstream tasks with limited labeled data~\cite{fewshot1, fewshot2}. In such a graph learning paradigm, pre-training and downstream graphs are often drawn from different domains and are rarely identically distributed. Differences in node attributes, graph topology, and underlying semantics induce distribution shifts across graphs~\cite{pretrain1, LEDA}. Consequently, most graph pre-training methods aim to mitigate these distribution discrepancies and learn more generalizable representations. 

\begin{figure}[t]
  \centering
  \includegraphics[width=\linewidth]{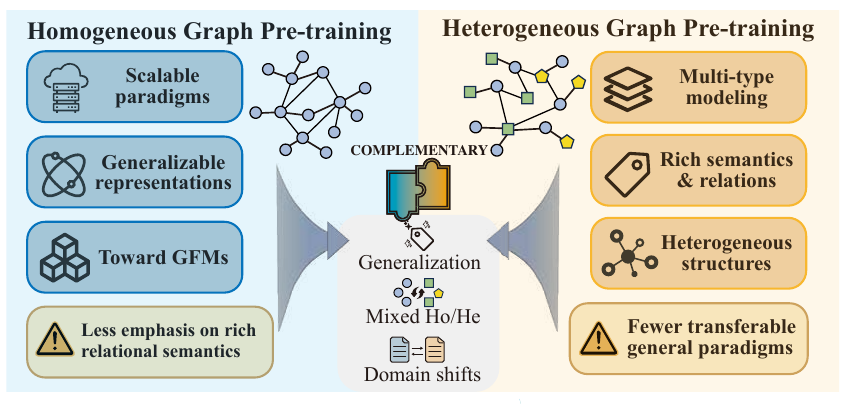}
  \caption{The fragmented progress of graph pre-training across homogeneous and heterogeneous graphs. Their complementary roles may be better suited to real-world mixed graph types and domain shifts.} 
  \label{fig:story}
\end{figure}

Conventionally, graph pre-training methods can be broadly divided into two categories based on graph types: homogeneous graph pre-training~\cite{SCOPE, fug, uniprompt} and heterogeneous graph pre-training~\cite{pretrain1, hetgpt, hgprompt, MUG}. Homogeneous graph pre-training primarily develops generalizable pre-training models and has become the principal driver of Graph Foundation Models (GFMs)~\cite{GFM1, GFM2}, yet it places less emphasis on the rich relational semantics inherent in the data. In contrast, heterogeneous graph pre-training predominantly emphasizes explicit modeling of multiple node and relation types~\cite{HINsurvey, metapath2vec}, focusing on capturing the intrinsic semantic structures of heterogeneous networks, whereas relatively few studies aim to learn transferable representations. Consequently, these two lines have advanced largely in parallel, with limited methodological overlap, resulting in uneven progress across graph types.

Real-world graphs are often neither purely homogeneous nor purely heterogeneous~\cite{PYG}. In practice, graph data from platforms or organizations typically involve both homogeneous interactions, e.g., social relationships~\cite{gnnSocialsurvey}, and heterogeneous relationships, e.g., user-item associations~\cite{HINsurvey2}, exhibiting substantial cross-domain discrepancies. In such mixed homogeneous-heterogeneous settings, existing pre-training methods cannot be directly applied to pre-train graphs across domains and graph types.

The above observations reveal a fundamental limitation in current graph learning: the fragmented progress in graph pre-training, largely separated between homogeneous and heterogeneous graphs, does not adequately address real-world requirements, in which mixed graph types and multi-domain distribution shifts are common. Importantly, our analysis suggests that the two lines of research are inherently complementary: homogeneous graph pre-training excels at learning scalable and generalizable representations, whereas heterogeneous graph pre-training is good at capturing various semantics and relationships. Their combination, therefore, offers substantial advantages and better fulfills real-world requirements. As illustrated in Fig.~\ref{fig:story}, this complementarity motivates us to explore joint pre-training across mixed graph types and domain shifts. In this work, we aim to develop a unified framework that supports joint multi-domain pre-training on mixed homogeneous and heterogeneous graphs, and to explore how such joint pre-training can better support downstream tasks. However, achieving such a goal is non-trivial, there are two main challenges:

\begin{itemize}[leftmargin=*, topsep=4pt, itemsep=0pt, parsep=0pt, beginpenalty=0, midpenalty=0]%
\item (\textbf{C1}) The incompatibility in network topology, both in terms of structural patterns and intrinsic semantics, arises from inherent differences between homogeneous and heterogeneous graphs, making it difficult to design a unified GNN encoder.
\item (\textbf{C2}) The fundamental differences between the two types of graphs further amplify the cross-domain shifts during joint pre-training, where domains may differ in graph types and feature semantics, leading to conflicting optimization signals and potential negative effects.
\end{itemize}

To address these challenges, we propose a unified multi-domain \textbf{G}raph \textbf{P}re-training method across \textbf{H}omogeneous and \textbf{H}eterogeneous graphs ($\mathbf{GPH^{2}}$). To unify encoding across homogeneous and heterogeneous graphs, we introduce \textbf{Unified Multi-View Graph Construction}, which transforms each graph into a set of graph views used as the encoder input. Specifically, for homogeneous graphs, we generate multiple augmented network topologies via data augmentation, i.e., by dropping edges~\cite{graphCL, GCA, GRACE}, whereas for heterogeneous graphs, we use meta-paths~\cite{metapath2vec, HAN} to create multiple topology patterns. In this way, GNN encoders remain agnostic to the graph type they process, whether homogeneous or heterogeneous. As a result, we can design a universal GNN to encode multi-view inputs and obtain the final representation by integrating across the outputs of these views, for both graph types.

To mitigate the impact of cross-domain distribution discrepancies, we introduce a \textbf{Domain-Specific Expert Encoding} process. Each expert is pre-trained on a single graph or several similar graphs, thereby maximizing knowledge extraction from the pre-training graphs while minimizing the potential negative effects of cross-domain distribution shift. When applied to downstream tasks, we propose an innovative \textbf{Task-Oriented Alignment and Fusion Strategy}, driven by the principle of \emph{"letting each expert focus on what it does best"}~\cite{graphMoE1}. In this strategy, a set of attention-based voting mechanisms are used to determine each expert's contribution to the task. For example, in a $k$-class classification task, we use $k$ voting mechanisms to compute class-wise attention scores for each expert, reflecting each expert's contributions to class discrimination. Through this design, the model adaptively integrates complementary strengths of different encoders, maximizing the effectiveness of multi-domain pre-training.

In summary, we make the following contributions:
\hangindent=1.0em
\hangafter=1
\subparagraph{1.} We first highlight the fragmented development of homogeneous and heterogeneous graph pre-training methods and empirically show that an appropriate mixture of homogeneous and heterogeneous graph pre-training can provide significant benefits for downstream tasks (see Section ~\ref{sec:motivational_study}).
\hangindent=1.0em
\hangafter=1
\subparagraph{2.} We introduce an innovative unified multi-domain graph pretraining method ($\text{GPH}^{2}$), bridging the gap between homogeneous and heterogeneous graph pre-training. By employing a Unified Multi-View Graph Construction technique, the method enables mixed graphs to be encoded with a common GNN architecture, without relying on graph-type-specific designs.
\hangindent=1.0em
\hangafter=1
\subparagraph{3.} We propose a Domain-Specific Expert Encoding process to mitigate the impact of cross-domain distribution discrepancies. Furthermore, we design a Task-Oriented Expert Fusion Strategy that adaptively integrates the outputs of multiple experts based on their discriminative strengths.
\hangindent=1.0em
\hangafter=1
\subparagraph{4.} We conduct comprehensive experiments on both homogeneous and heterogeneous graphs, demonstrating that $\text{GPH}^{2}$ enables stable transfer across graph types and domains, and systematically validate the effectiveness of each component in $\text{GPH}^{2}$.
\hangindent=1.0em
\hangafter=1

\section{Problem Definition and Motivational Study}
\subsection{Problem Definition}
We consider a multi-domain graph pre-training and transfer-learning setting involving both homogeneous and heterogeneous graphs. A homogeneous graph is defined as ${{\mathcal{G}}^{{}}}=(\mathcal{V},\mathcal{E},\mathbf{X})$, where all nodes $v\in \mathcal{V}$ share the same node type, all edges $e\in \mathcal{E}$ share the same relation type, and $\mathbf{X}\in {{\mathbb{R}}^{|\mathcal{V}|\times d}}$ denotes node features. 
A heterogeneous graph is defined as ${{\mathcal{G}}^{{}}}=(\mathcal{V},\mathcal{E},{{\mathcal{T}}_{v}},{{\mathcal{T}}_{e}},\mathbf{X}),$, where nodes and edges are associated with type mappings ${{\mathcal{T}}_{v}}:\mathcal{V}\to \mathcal{A}$, ${{\mathcal{T}}_{e}}:\mathcal{E}\to \mathcal{R}$ with types $\left| \mathcal{A} \right|+\left| \mathcal{R} \right|>1$.

For heterogeneous graph, a meta-path $P$ is a path of node and edge types, denoted as 
${{A}_{1}}\xrightarrow{{{R}_{1}}}{{A}_{2}}\xrightarrow{{{R}_{2}}}\cdots \xrightarrow{{{R}_{l}}}{{A}_{l+1}}$
(abbreviated as ${{A}_{1}}{{A}_{2}}\cdots {{A}_{l+1}}$), where ${{A}_{i}}\in \mathcal{A}$ and ${{R}_{i}}\in \mathcal{R}$.

Let ${{\mathcal{D}}}=\{{{\mathcal{G}}_{1}},{{\mathcal{G}}_{2}},\ldots ,{{\mathcal{G}}_{M}}\}$ denote a set of graph set from multiple domains, where each ${{\mathcal{G}}_{i}}$ may be either homogeneous or heterogeneous. The process of multi-domain pre-training is
\begin{equation}\label{eq:definition_pretrain_encoder}
	{{f}_{i}}:{{\mathcal{G}}_{i}}\to {{\mathbf{H}}_{i}},\quad i=1,\ldots ,\;M,
\end{equation}
where ${{f}_{i}}$ is parameterized by ${{\theta }_{i}}$, which can be implemented as a shared encoder or as a set of domain-specific encoders, and ${\mathbf{H}}_{i}$ is the representation of graph $\mathcal{G}_i$. We focus on the latter paradigm and pre-train an independent encoder for each domain to better preserve domain-specific structural and semantic characteristics:
\begin{equation}\label{eq:definition_pretrain_loss}
	\underset{{{\theta }_{i}}}{\mathop{\min }}\,\mathcal{L}_{pre}^{(i)}({{f}_{i}}({{\mathcal{G}}_{i}})),\quad i=1,\ldots ,\;M.
\end{equation}
where $\mathcal{L}_{pre}$ can be arbitrary self-supervised learning objectives. 

\begin{figure}[htbp]
  \centering
  \includegraphics[width=\linewidth]{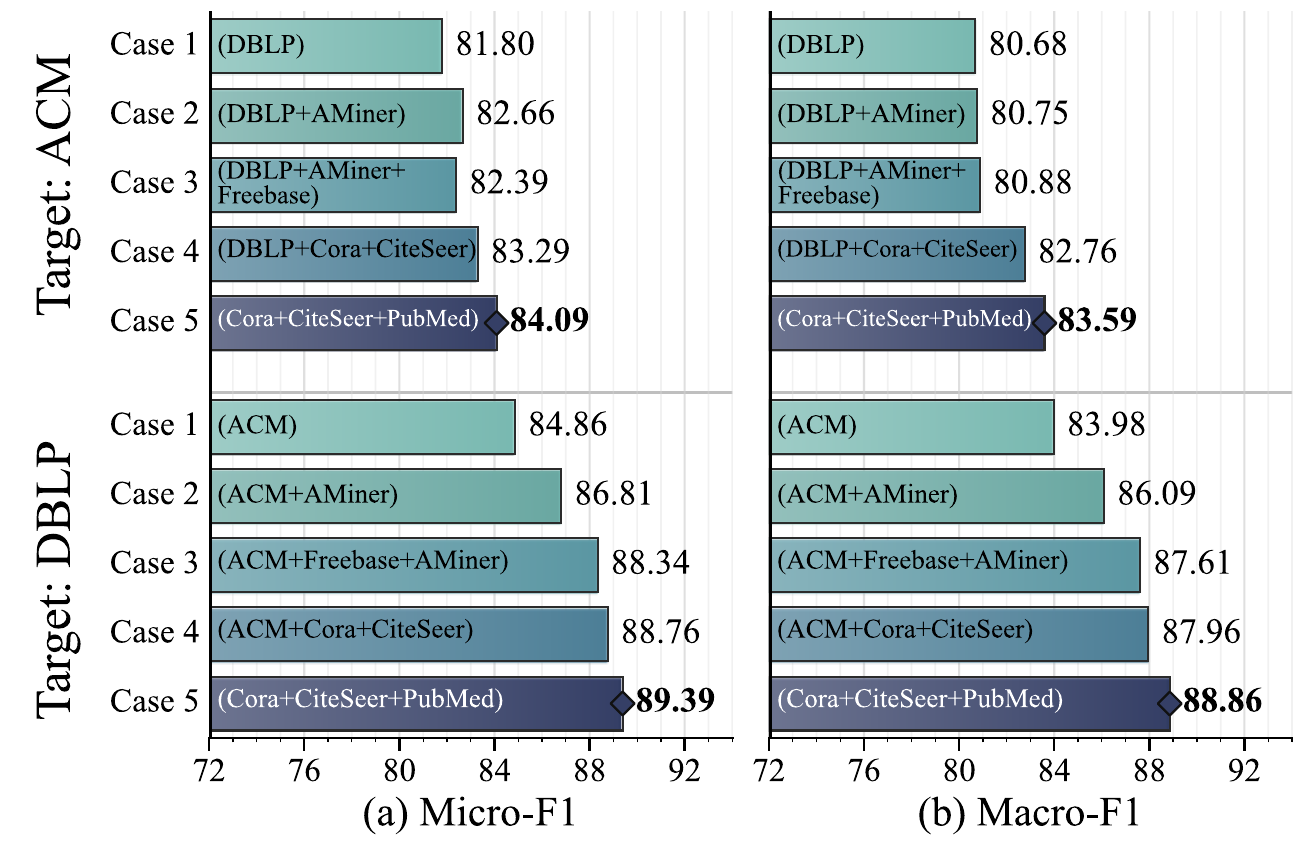}
  \caption{Performance of 3-shot node classification under different pre-training graphs on ACM and DBLP.}
  \label{fig:motivational_study}
\end{figure}

\subsection{Motivational Study}\label{sec:motivational_study}
To appreciate how different pre-training graphs influence downstream performance, we conduct $3$-shot node classification experiments on two widely used public benchmarks, i.e., ACM and DBLP, as downstream tasks (the model and experiment settings are detailed in Section ~\ref{sec:overview} and Section ~\ref{sec:experiment_setting}). The pre-training graphs consist of three homogeneous citation networks (Cora~\cite{Cora}, CiteSeer~\cite{citeseerPubmed}, and PubMed~\cite{citeseerPubmed}) and four heterogeneous networks (ACM, DBLP, Aminer, and Freebase)~\cite{HeCo}. Among them, Cora, CiteSeer, PubMed, ACM, and DBLP are widely regarded as high-quality datasets, as a broad range of models achieve strong performance on traditional node classification tasks~\cite{GCN,GAT,HAN,MAGNN}. From a pre-training perspective, this suggests that they may provide richer structural and semantic signals for downstream tasks.

From the results in Figure ~\ref{fig:motivational_study}, we have two key observations:
\subparagraph{\textbf{Obs.1: Increasing Pre-training Domain Diversity Improves Downstream Performance.}} From Case 1-3, as the pre-training graphs become progressively more diverse, downstream performance consistently improves. This trend indicates that multi-domain pre-training enables richer and more transferable representations, even when pre-training and downstream domains are not closely related.
\subparagraph{\textbf{Obs.2: Incorporating Homogeneous Graphs Can Benefit Heterogeneous Downstream Tasks.}} According to Case 3-5, despite structural differences between homogeneous and heterogeneous graphs, incorporating homogeneous citation graphs into the pre-training consistently improves performance on heterogeneous downstream tasks. Notably, in Case 5, pre-training only using homogeneous graphs even achieves the best downstream performance. This observation can be plausibly explained by the high domain similarity between Cora, CiteSeer, PubMed and the downstream ACM and DBLP datasets, as they all originate from citation networks and share similar structural and semantic characteristics.

These observations indicate that homogeneous and heterogeneous graphs are not mutually exclusive in the pre-training stage. Instead, an appropriate combination of both can substantially enhance downstream performance, with the key lying in the similarity between the pre-training graphs and the downstream graph rather than in their graph types. Overall, the empirical evidence strongly suggests that joint pre-training over homogeneous and heterogeneous graphs is both feasible and beneficial. This observation underscores the need for a unified, multi-domain pre-training framework that bridges homogeneous and heterogeneous graphs.

\section{Method}\label{sec:method}
In this section, we present $\text{GPH}^{2}$, a unified multi-domain graph pre-training method  for both homogeneous and heterogeneous graphs. To address two challenges outlined in Section ~\ref{sec:intro}, we design $\text{GPH}^{2}$ with two key components: (1) a Unified Multi-View Graph Construction that reformulates both homogeneous and heterogeneous graphs into a unified graph form, and (2) a Domain-Specific Expert Encoding that captures domain-specific knowledge during pre-training and integrates them effectively at downstream stage.

\begin{figure*}[t]
  \centering
  \includegraphics[width=\linewidth]{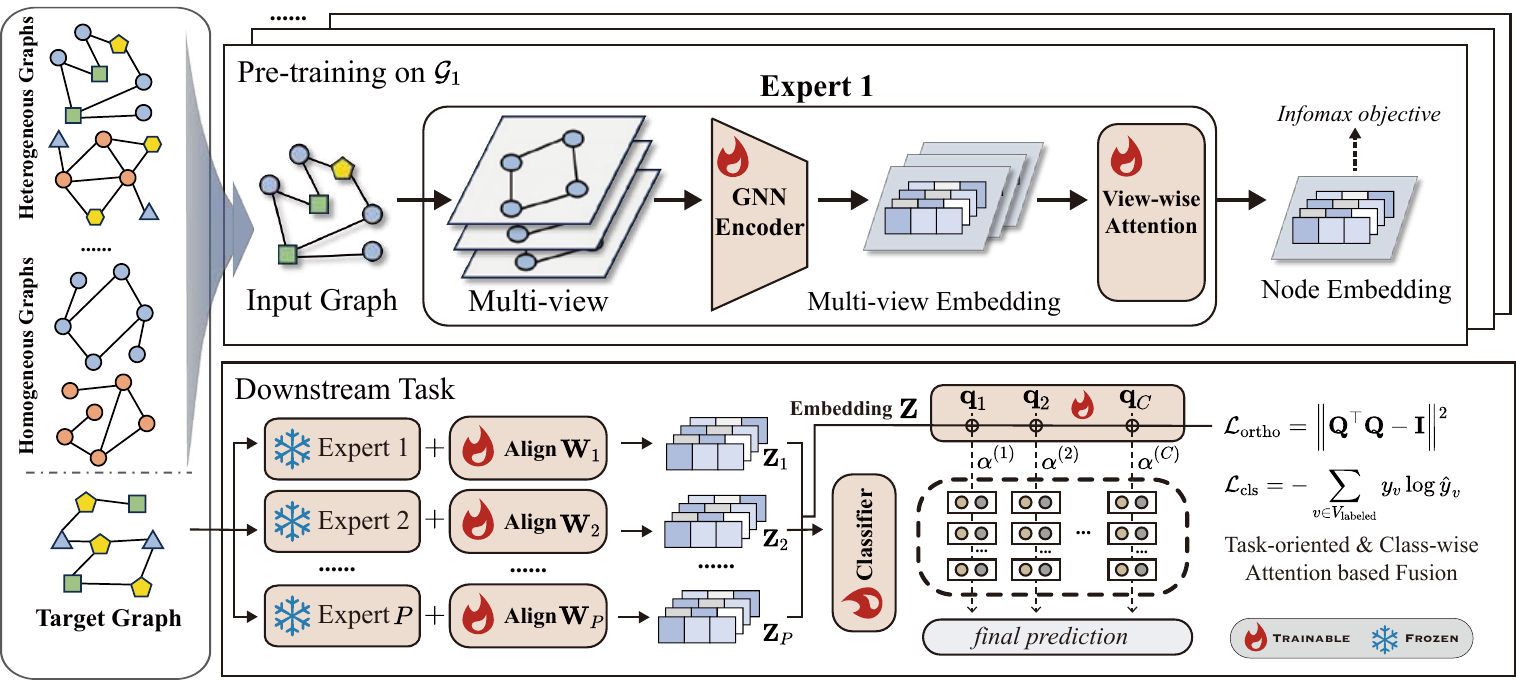}
  \caption{The workflow of $\text{GPH}^{2}$. $\text{GPH}^{2}$ is designed for learning across mix graphs (left). It consists of two stages: multi-domain pre-training (top) and downstream task adaptation (bottom). In the pre-training stage, each source graph is independently processed by a domain-specific expert, and in the downstream stage, all pre-trained experts are transferred for adaptation.}
  \label{fig:overview}
\end{figure*}

\subsection{Overview}\label{sec:overview}
The overall framework of $\text{GPH}^{2}$ is illustrated in Figure ~\ref{fig:overview}. The workflow is divided into two stages: a pre-training stage and a downstream task-adaptation stage.

In the pre-training stage, both homogeneous and heterogeneous graphs are first transformed into a unified multi-view graph form through the proposed Unified Multi-View Graph Construction module. This module converts multiple graphs into a unified input (i.e., a multi-view graph), enabling arbitrary graph neural networks (GNNs) to support both homogeneous and heterogeneous graph encodings. Subsequently, multiple GNN encoders with non-shared parameters are trained independently in a self-supervised manner, where each encoder is associated with a graph. This allows each encoder to focus on learning domain-specific representations without being affected by other domains.

In the downstream stage, the target graph is first reformulated as a multi-view graph. Each pre-trained GNN encoder then acts as an expert and independently produces representations of the target graph. These representations are subsequently aligned into a shared space. Finally, a task-oriented attention module is employed to adaptively estimate each expert's contribution and aggregate their predictions to produce the final decision. It is worth emphasizing that $\text{GPH}^{2}$ is an open and paradigm-level pre-training framework. In principle, arbitrary GNNs and self-supervised pre-training objectives can be seamlessly integrated into our framework.

\subsection{Unified Multi-View Graph Construction}
Unified Multi-View Graph Construction reformulates graphs from different domains into a common graph form, abstracting away incompatible semantics in node features and network topologies. Importantly, it is entirely training-free: rather than learning complex transformations, it provides a unified input for GNN encoders.

In a homogeneous graph, all nodes and edges share the same types. To capture diverse structural patterns, we generate multiple graph views via topology-level data augmentation. Specifically, we adopt edge dropping~\cite{graphCL} to construct an augmented edge set, i.e.,
\begin{equation}
	{{\mathcal{E}}^{(p)}}\subseteq \mathcal{E},\quad k=1,\ldots ,P \;,
\end{equation}
where the augmented edge set ${{\mathcal{E}}^{(k)}}$ is obtained by randomly removing some edges from $\mathcal{E}$.

In a heterogeneous graph, meta-paths explicitly characterize semantically meaningful connectivity patterns. Given a set of predefined meta-paths $\{{{\pi }_{1}},\ldots ,{{\pi }_{P}}\}$, we rewire the graph based on meta-path connectivity~\cite{HAN}. Formally, the edge set induced by the meta-path ${{\pi }_{p}}$ is defined as: 
\begin{equation}
	{{\mathcal{E}}^{(p)}}=\{(u,v)\mid \text{ }u\text{ and }v\text{ are connected by a }{{\pi }_{p}}\},
\end{equation}
The multi-view graph effectively captures the structure of the original graph. Moreover, it enhances the expressiveness of topological semantics, particularly in heterogeneous graphs where meta-paths explicitly refine the extraction of meaningful relations.

The above construction reformulates both homogeneous and heterogeneous graphs into a unified multi-view form:
\begin{equation}
	\mathcal{G}\ \to \ \{{{\mathcal{G}}^{(1)}},{{\mathcal{G}}^{(2)}},\ldots ,{{\mathcal{G}}^{(P)}}\},
\end{equation}
where each view ${\mathcal{G}}^{(p)}$ is defined on the same node set with a unified feature space, while differing only in their edge sets. In principle, any GNN model can be employed as the encoder to process these views, so the representation of a view can be expressed as
\begin{equation}
	\mathbf{H}_{{}}^{(p)}=\text{GNN}(\mathbf{X},A_{{}}^{(p)};\theta ),
\end{equation}
where ${{A}^{(p)}}$ denotes the adjacency matrix corresponding to ${{\mathcal{E}}^{(p)}}$, and $\theta$ represents encoder parameters shared across all views. We further aggregate the view-specific representations to obtain final node representations via an attention-based aggregation, i.e.,
\begin{equation}
    \alpha _{p}^{{}}=\frac{\exp \left( {{\mathbf{q}}^{\top }}\mathbf{h}_{v}^{(p)} \right)}{\sum\nolimits_{p'=1}^{P}{\exp\left( {{\mathbf{q}}^{\top }}\mathbf{h}_{v}^{({p}')} \right)}},
\end{equation}
where $\mathbf{h}_{v}^{(p)}$ denotes the representation of node $v$ in view $p$, and $\mathbf{q}$ is a learnable attention vector.

\subsection{Domain-Specific Expert Encoding}
In multi-domain graph pre-training, graphs from different domains typically exhibit significant distributional  discrepancies~\cite{SAMGPT, uniprompt}. These discrepancies arise from variations in graph size, node attributes, and, more critically, topological structural patterns. When homogeneous and heterogeneous graphs are considered jointly, this issue becomes more pronounced. For example, homogeneous citation networks typically follow power-law degree distributions~\cite{powerLaw} and exhibit relatively uniform relational semantics, whereas heterogeneous citation networks such as DBLP contain multiple node and relation types with highly imbalanced topology patterns across types (e.g., average node degree in DBLP: $4.8$ for \emph{Author}, $8.4$ for \emph{Paper}, $11.1$ for \emph{Term} and $716.4$ for \emph{Conference}). As a result, a single shared GNN is often insufficient to capture these diverse structural characteristics simultaneously and may yield suboptimal representations when trained across multiple domains.

Instead of enforcing parameter sharing across domains, we introduce a domain-specific expert pre-training process and assign one expert encoder to one graph or several similar graphs. Each expert is trained independently and in parallel, allowing it to specialize in modeling the structural and semantic properties of its own domain without being affected by distribution shifts from others, providing a more reliable downstream adaptation and expert selection.

Formally, we associate each graph ${{\mathcal{G}}_{i}}\in \{{{\mathcal{G}}_{1}},{{\mathcal{G}}_{2}},\ldots ,{{\mathcal{G}}_{M}}\}$ with an encoder${{f}_{i}}(\cdot ;{{\theta }_{i}},{{\mathbf{q}}_{i}})\in \{f_{1}, f_{2},\ldots , f_{M} \}$, where each GNN parameters ${{\theta }_{i}}$ and attention vector $\mathbf{q}_i$ are not shared across experts. Each expert encoder processes its corresponding graph independently to obtain node representations, which is as follows:

\begin{equation}
	{{\mathbf{H}}_{i}}={{f}_{i}}({{\mathcal{G}}_{i}};{{\theta }_{i}},{{\mathbf{q}}_{i}}).
\end{equation}
The pre-training of different experts is fully decoupled and can be conducted in parallel, ensuring that each encoder maximally captures domain-specific knowledge.

In $\text{GPH}^{2}$, any self-supervised learning objective can be employed to train the expert encoders. In this work, we follow the information maximization (Infomax) principle~\cite{DGI, DMGI}, which aims to maximize the mutual information between local node representations and global graph-level summaries.

Specifically, for each expert ${{f}_{i}}$, we adopt a DGI-style objective~\cite{DGI}. Let ${{\mathbf{H}}_{i}}$ be the node representation of $\mathcal{G}_i$ and ${{\mathbf{s}}_{i}}$ be the corresponding graph summary vector:
\begin{equation}
	{{\mathbf{s}}_{i}}=\text{Readout}({{\mathbf{H}}_{i}}),
\end{equation}
where Readout denotes a permutation-invariant aggregation function (e.g., mean pooling). Negative samples are generated by corrupting the input graph (e.g., feature shuffling), yielding corrupted node representations ${{\widetilde{\mathbf{H}}}_{i}}$. The Infomax objective for expert ${{f}_{i}}$ is defined as:
\begin{equation}
	{{\mathcal{L}}_{i}}=-\sum\limits_{v\in {{V}_{i}}}{\left[ \log \sigma \left( \mathbf{h}_{i,v}^{\top }{{\mathbf{s}}_{i}} \right)+\log \left( 1-\sigma \left( \widetilde{\mathbf{h}}_{i,v}^{\top }{{\mathbf{s}}_{i}} \right) \right) \right]},
\end{equation}
where $\sigma (\cdot )$ is the sigmoid function.

\subsection{Task-oriented Expert Fusion}
Given a graph $\mathcal{G}$ for a downstream task, we first reformulate it into a unified multi-view form and then feed it into all pre-trained expert encoders $\{{{f}_{i}}\}_{i=1}^{P}$, whose parameters are frozen during downstream training, which is as follows:
\begin{equation}
	\mathbf{H}_{i}^{{}}={{f}_{i}}(\mathcal{G}),
\end{equation}
where ${{\mathbf{H}}_{i}}$ is the representation produced by ${{f}_{i}}$. Since the experts are pre-trained on different domains, their output representations are generally not directly comparable when applied to the downstream task, as they encode domain-specific biases induced by cross-domain discrepancies. Here, we introduce the expert-specific alignment to map all the representations into a shared space:
\begin{equation}\label{eq:alignment}
	{{\mathbf{Z}}_{i}}={{\mathbf{W}}_{i}}\mathbf{H}_{i}^{{}},
\end{equation}
where ${{\mathbf{W}}_{i}}$ is an expert-specific linear projection matrix, and ${{\mathbf{Z}}_{i}}$ is the aligned representation in the shared embedding space.

For a $C$-class classification task, we produce $C$-dimensional logit vectors for node $v$ based on aligned representation $\mathbf{Z}_i$, i.e.,
\begin{equation}
	{{\ell }_{i,v}}=\phi ({{\mathbf{z}}_{i,v}})\in {{\mathbb{R}}^{C}},
\end{equation}
where $\phi (\cdot )$ is a classifier shared across all representation $\{{{\mathbf{Z}}_{i}}\}_{i=1}^{P}$, $\mathbf{z}_{i,v}$ is the embedding of node $v$ in $\mathbf{Z}_i$. Here, the $c$-th element $\ell _{i,v}^{(c)}$ represents $i$-th expert's confidence score for assigning node $v$ to class $c$. The final prediction for node $v$ should be obtained by fusing all the logit vectors of node $v$, i.e., $\{\ell _{i,v}^{{}}\}_{i=1}^{P}$.

A naive fusion strategy, e.g., uniform averaging, may yield suboptimal predictions because it implicitly assumes that all experts contribute equally across all classes. However, in practice, experts often exhibit varying discriminative strengths across classes. For example, an expert pre-trained on citation networks may be more effective at identifying research-area-related classes, while another expert pre-trained on e-commerce networks may excel at capturing product- or transaction-related semantics.

To capture such behavior, we adopt a task-oriented, class-wise attention voting mechanism. We first initialize a set of learnable class-specific attention vectors:
\begin{equation}\label{eq:attention_vectors}
	\mathcal{Q}=\{{{\mathbf{q}}_{1}},{{\mathbf{q}}_{2}},\ldots ,{{\mathbf{q}}_{C}}\},\quad {{\mathbf{q}}_{c}}\in {{\mathbb{R}}^{d}}.
\end{equation}
For each class $c$, we compute the contribution of expert ${{f}_{i}}$ by measuring the similarity between the aligned representation $\mathbf{Z}_i$ and the corresponding attention vector, which can be expressed as: 
\begin{equation}
	\alpha _{i}^{(c)}=\frac{\exp (\mathbf{Z}_{i}^{\top }{{\mathbf{q}}_{c}})}{\sum\nolimits_{j=1}^{P}{\exp(\mathbf{Z}_{j}^{\top }{{\mathbf{q}}_{c}})}}.
\end{equation}

\begin{table*}[!t]
\centering
\caption{Experiments results on few-shot node classification for homogeneous graphs. We report the average performance for 10 repetitions. The best results are highlighted in bold. \textit{Avg.} represents the mean performance of all datasets. }
\label{tab:fewshot_ho}
\begin{tabular}{c!{\vrule width 0.5pt}c!{\vrule width 0.5pt}c!{\vrule width 0.5pt}cccccc!{\vrule width 0.5pt}c}
\toprule
Dataset & Metric & Labels & DGI & GRACE & GraphMAE & GCOPE & MDGPT & MDGFM & $\mathbf{GPH^{2}}$ \\
\midrule
\multirow{4}{*}{Cora}       & \multirow{2}{*}{Micro-F1}  & 3  & 50.12$\pm$6.01 & 59.09$\pm$4.71 & 51.39$\pm$5.66 & 59.64$\pm$6.05 & 43.15$\pm$4.84 & 51.01$\pm$4.62 & \textbf{63.59}$\pm$3.54 \\
                            &                            & 5  & 53.49$\pm$3.66 & 67.07$\pm$2.40 & 57.70$\pm$4.47 & 67.20$\pm$5.13 & 50.71$\pm$4.00 & 57.00$\pm$3.76 & \textbf{68.58}$\pm$3.74 \\
                            & \multirow{2}{*}{Macro-F1}  & 3  & 48.19$\pm$6.54 & 58.30$\pm$4.95 & 48.05$\pm$5.34 & 60.32$\pm$6.06 & 43.99$\pm$5.31 & 50.60$\pm$4.08 & \textbf{62.10}$\pm$4.19 \\
                            &                            & 5  & 51.57$\pm$3.97 & 66.85$\pm$2.48 & 54.43$\pm$4.25 & 66.44$\pm$4.98 & 51.43$\pm$4.46 & 57.26$\pm$3.40 & \textbf{67.87}$\pm$3.35 \\
\midrule
\multirow{4}{*}{CiteSeer}   & \multirow{2}{*}{Micro-F1}  & 3  & 43.65$\pm$6.27 & 42.99$\pm$4.10 & 49.24$\pm$6.93 & \textbf{58.78}$\pm$1.19 & 42.54$\pm$5.57 & 38.27$\pm$4.95 & 55.15$\pm$3.56 \\
                            &                            & 5  & 48.78$\pm$4.99 & 47.87$\pm$3.73 & 56.14$\pm$2.64 & \textbf{69.03}$\pm$0.33 & 48.95$\pm$5.96 & 43.43$\pm$3.75 & 60.02$\pm$4.38 \\
                            & \multirow{2}{*}{Macro-F1}  & 3  & 40.87$\pm$5.82 & 39.92$\pm$3.97 & 46.02$\pm$6.17 & \textbf{55.02}$\pm$1.69 & 38.88$\pm$5.14 & 36.31$\pm$4.61 & 51.08$\pm$3.21 \\
                            &                            & 5  & 45.51$\pm$4.70 & 45.04$\pm$3.70 & 52.50$\pm$2.81 & \textbf{66.50}$\pm$0.66 & 45.16$\pm$5.66 & 41.22$\pm$3.26 & 56.78$\pm$3.90 \\
\midrule
\multirow{4}{*}{PubMed}     & \multirow{2}{*}{Micro-F1}  & 3  & 56.33$\pm$4.36 & 57.43$\pm$3.42 & 58.69$\pm$8.24 & 54.65$\pm$1.17 & OOM            &  OOM           & \textbf{62.96}$\pm$8.78 \\
                            &                            & 5  & 64.47$\pm$4.00 & 64.33$\pm$3.05 & 65.53$\pm$4.67 & 63.04$\pm$0.52 & OOM            &  OOM           & \textbf{67.76}$\pm$3.56 \\
                            & \multirow{2}{*}{Macro-F1}  & 3  & 56.68$\pm$4.39 & 56.70$\pm$3.99 & 58.68$\pm$8.46 & 51.85$\pm$1.25 & OOM            &  OOM           & \textbf{62.11}$\pm$9.88 \\
                            &                            & 5  & 64.63$\pm$4.17 & 64.28$\pm$3.00 & 64.95$\pm$4.71 & 62.88$\pm$0.79 & OOM            &  OOM           & \textbf{67.30}$\pm$4.33 \\
\midrule
\multirow{4}{*}{Photo}      & \multirow{2}{*}{Micro-F1}  & 3  & 62.42$\pm$6.51 & 67.30$\pm$7.41 & 71.22$\pm$9.09 & 64.46$\pm$1.87 & 76.07$\pm$5.21 & 67.63$\pm$4.75 & \textbf{77.22}$\pm$2.71 \\
                            &                            & 5  & 67.83$\pm$4.12 & 78.88$\pm$4.05 & 77.05$\pm$5.10 & 66.36$\pm$1.86 & 80.14$\pm$3.83 & 72.33$\pm$4.53 & \textbf{82.32}$\pm$2.22 \\
                            & \multirow{2}{*}{Macro-F1}  & 3  & 60.29$\pm$6.15 & 67.60$\pm$6.79 & 70.36$\pm$8.49 & 64.61$\pm$2.00 & 75.01$\pm$4.71 & 66.19$\pm$4.35 & \textbf{75.30}$\pm$2.81 \\
                            &                            & 5  & 65.65$\pm$4.46 & 78.17$\pm$3.99 & 76.45$\pm$4.62 & 65.46$\pm$1.50 & 78.98$\pm$3.29 & 70.80$\pm$3.84 & \textbf{80.22}$\pm$1.81 \\
\midrule
\multirow{4}{*}{Computer}   & \multirow{2}{*}{Micro-F1}  & 3  & 52.89$\pm$6.06 & 60.69$\pm$6.81 & 56.58$\pm$8.36 & 56.90$\pm$1.69 & OOM            & 53.12$\pm$6.18 & \textbf{61.25}$\pm$7.17 \\
                            &                            & 5  & 57.49$\pm$4.20 & 69.41$\pm$3.45 & 65.02$\pm$5.19 & 61.10$\pm$1.80 & OOM            & 60.58$\pm$4.71 & \textbf{69.62}$\pm$2.61 \\
                            & \multirow{2}{*}{Macro-F1}  & 3  & 50.28$\pm$5.78 & \textbf{63.50}$\pm$6.60 & 54.17$\pm$4.80 & 54.51$\pm$1.59 & OOM            & 52.58$\pm$5.66 & 61.40$\pm$4.50 \\
                            &                            & 5  & 54.80$\pm$4.34 & \textbf{71.18}$\pm$3.45 & 60.62$\pm$5.53 & 60.94$\pm$0.85 & OOM            & 59.47$\pm$4.16 & 68.97$\pm$2.80 \\
\midrule
\multicolumn{2}{r}{Avg. \quad\quad}								&  & 54.80		   & 61.33			& 59.74			 & 61.48		  & 56.25		   & 54.86			& \textbf{66.07}		  \\
\bottomrule
\hline
\end{tabular}
\end{table*}

\begin{table*}[!t]
\centering
\caption{Experiments results on few-shot node classification heterogeneous graphs. We report the average performance for 10 repetitions. The best results are highlighted in bold. \textit{Avg.} represents the mean performance of all datasets.}
\label{tab:fewshot_he}
\begin{tabular}{c!{\vrule width 0.5pt}c!{\vrule width 0.5pt}c!{\vrule width 0.5pt}cccccc!{\vrule width 0.5pt}c}
\toprule
Dataset & Metric & Labels & DMGI & HeCo & HGMAE & HERO & HGPrompt & HetGPT & $\mathbf{GPH^{2}}$ \\
\midrule
\multirow{4}{*}{ACM}        & \multirow{2}{*}{Micro-F1}  & 3  & 72.81$\pm$7.66 & 74.28$\pm$10.5 & 74.93$\pm$11.4  & 56.69$\pm$5.58 & 72.97$\pm$8.08 & 61.10$\pm$6.03 & \textbf{82.39}$\pm$2.86 \\
                            &                            & 5  & 74.23$\pm$7.81 & 78.76$\pm$6.97 & 75.08$\pm$7.96  & 65.11$\pm$4.57 & 70.44$\pm$9.37 & 73.16$\pm$8.62 & \textbf{82.46}$\pm$8.03 \\
                            & \multirow{2}{*}{Macro-F1}  & 3  & 69.18$\pm$9.56 & 71.41$\pm$11.9 & 72.66$\pm$10.7  & 56.77$\pm$5.74 & 72.36$\pm$6.83 & 55.83$\pm$7.55 & \textbf{80.88}$\pm$4.59 \\
                            &                            & 5  & 74.92$\pm$7.75 & 79.17$\pm$5.97 & 75.44$\pm$7.74  & 65.84$\pm$4.65 & 70.43$\pm$7.64 & 70.18$\pm$13.6 & \textbf{82.77}$\pm$7.28 \\
\midrule
\multirow{4}{*}{DBLP}       & \multirow{2}{*}{Micro-F1}  & 3  & 57.43$\pm$9.43 & 77.55$\pm$11.7 & 82.16$\pm$6.14  & 87.66$\pm$0.86 & 87.27$\pm$3.01 & 72.08$\pm$8.69 & \textbf{88.34}$\pm$2.58 \\
                            &                            & 5  & 68.35$\pm$4.68 & 83.86$\pm$5.57 & 85.54$\pm$2.72  & 88.14$\pm$1.27 & 88.07$\pm$2.23 & 70.83$\pm$6.05 & \textbf{90.36}$\pm$1.13 \\
                            & \multirow{2}{*}{Macro-F1}  & 3  & 56.13$\pm$9.10 & 76.07$\pm$12.0 & 80.71$\pm$7.03  & 86.91$\pm$0.97 & 86.03$\pm$3.14 & 70.63$\pm$8.65 & \textbf{87.61}$\pm$2.52 \\
                            &                            & 5  & 67.49$\pm$4.75 & 83.31$\pm$5.11 & 84.90$\pm$2.66  & 87.49$\pm$1.28 & 87.05$\pm$2.33 & 68.77$\pm$6.30 & \textbf{89.66}$\pm$1.17 \\
\midrule
\multirow{4}{*}{Aminer}     & \multirow{2}{*}{Micro-F1}  & 3  & 29.93$\pm$5.36 & 29.93$\pm$5.45 & 34.91$\pm$6.80  & 26.87$\pm$5.64 & 30.20$\pm$5.08 & 37.12$\pm$11.6 & \textbf{45.98}$\pm$8.34 \\
                            &                            & 5  & 26.73$\pm$5.83 & 29.24$\pm$5.67 & 35.88$\pm$6.07  & 29.50$\pm$6.96 & 34.01$\pm$5.37 & \textbf{46.57}$\pm$11.1 & 37.57$\pm$8.90 \\
                            & \multirow{2}{*}{Macro-F1}  & 3  & 28.73$\pm$1.87 & 24.48$\pm$1.98 & 28.33$\pm$3.47  & 24.14$\pm$5.58 & 25.99$\pm$2.85 & 20.09$\pm$6.26 & \textbf{33.88}$\pm$6.09 \\
                            &                            & 5  & 24.02$\pm$3.02 & 24.58$\pm$2.96 & 30.60$\pm$3.40  & 24.08$\pm$9.01 & 28.83$\pm$3.01 & 19.94$\pm$3.85 & \textbf{33.55}$\pm$6.64 \\
\midrule
\multirow{4}{*}{Freebase}   & \multirow{2}{*}{Micro-F1}  & 3  & 33.92$\pm$3.07 & 40.67$\pm$2.51 & 38.55$\pm$3.50  & 35.33$\pm$7.56 & 41.97$\pm$3.40 & 43.86$\pm$2.32 & \textbf{45.34}$\pm$3.45 \\
                            &                            & 5  & 35.22$\pm$3.01 & 42.77$\pm$3.15 & 37.39$\pm$4.50  & 33.67$\pm$8.16 & 43.49$\pm$3.26 & 42.22$\pm$3.45 & \textbf{47.88}$\pm$6.93 \\
                            & \multirow{2}{*}{Macro-F1}  & 3  & 31.95$\pm$2.60 & 35.43$\pm$3.25 & 34.79$\pm$3.58  & 31.39$\pm$7.80 & 37.55$\pm$3.12 & 27.57$\pm$8.21 & \textbf{39.73}$\pm$4.68 \\
                            &                            & 5  & 33.37$\pm$2.32 & 37.99$\pm$3.12 & 35.15$\pm$3.62  & 30.35$\pm$8.38 & 40.78$\pm$3.23 & 29.03$\pm$6.54 & \textbf{44.78}$\pm$5.47 \\
\midrule
\multicolumn{2}{r}{Avg. \quad\quad}								&  & 49.02		   & 55.59			& 56.67			  & 51.87		   & 57.34			& 50.56			 & \textbf{63.32}		   \\			
\bottomrule
\end{tabular}
\end{table*}

The node $v$’s final prediction for class $c$ is obtained by aggregating expert logits using the corresponding attention weights:
\begin{equation}\label{eq:attention_fusion}
	\ell _{v}^{(c)}=\sum\limits_{i=1}^{P}{\alpha _{i}^{(c)}}\ell _{i,v}^{(c)}.
\end{equation}

This class-wise fusion mechanism allows the model to adaptively select the most informative experts for each class, rather than enforcing a uniform contribution across experts. To encourage diversity and specialization among class-specific attention vectors, we introduce orthogonality regularization,
\begin{equation}\label{eq:orthogonality}
	{{\mathcal{L}}_{\text{ortho}}}=\left\| {{\mathbf{Q}}^{\top }}\mathbf{Q}-\mathbf{I} \right\|_{{}}^{2},
\end{equation}
which penalizes correlation between different attention vectors. The final downstream optimization objective is:
\begin{equation}
	{{\mathcal{L}}_{\text{down}}}={{\mathcal{L}}_{\text{cls}}}+{{\mathcal{L}}_{\text{ortho}}},
\end{equation}
where $ {{\mathcal{L}}_{\text{cls}}}$ is the cross-entropy loss.

\section{Experiment}
\subsection{Experiment Settings}\label{sec:experiment_setting}
We evaluate $\text{GPH}^{2}$ on 5 homogeneous graphs and 4 heterogeneous graphs, and compare it with 6 homogeneous graph baselines and 6 heterogeneous graph baselines, respectively. To the best of our knowledge, no existing graph pre-training method explicitly supports mixtures of homogeneous and heterogeneous graphs. Therefore, we compare $\text{GPH}^{2}$ with homogeneous and heterogeneous baselines separately. More experiment settings can be found in Appendix, and the code is available at https://github.com/hedongxiao-tju/GPH-2.

\subsection{Performance Evaluation}
We evaluated $\text{GPH}^{2}$ on 3- and 5-shot node classification tasks. Due to the limitations of baselines, we first conducted comparisons under matched graph-type settings, i.e., homogeneous-to-homogeneous and heterogeneous-to-heterogeneous pre-training and transfer. For each downstream graph, all remaining graphs are used for pre-training. The corresponding results are reported in Table ~\ref{tab:fewshot_ho} and ~\ref{tab:fewshot_he}, respectively.
Overall, $\text{GPH}^{2}$ consistently achieves the best performance across both homogeneous and heterogeneous few-shot node classification tasks. On homogeneous graphs, it outperformed the strongest baseline, SCOPE, by an average of $4.59\%$, and on heterogeneous graphs it exceeded HGPrompt by $5.98\%$. These consistent improvements demonstrate the effectiveness and generality of $\text{GPH}^{2}$. Second, although several baselines have their own downstream adaptation methods, they made a slight improvement compared to standard self-supervised GNNs with a simple classifier(e.g., SCOPE vs. GRACE: $+0.15\%$, HGPrompt vs. HeCo: $+0.67\%$). In contrast, $\text{GPH}^{2}$ yielded substantially improvement, indicating that domain-specific expert encoding combined with task-oriented expert fusion more effectively mitigates cross-domain distribution shifts. Third, almost all baseline methods exhibited limited robustness across datasets. For instance, although SCOPE performs strongly on CiteSeer, it shows performance degradation on Photo and Computer compared with GRACE, revealing poor generalization across domains. In contrast, $\text{GPH}^{2}$ maintains stable and strong performance across all graphs, further validating its robustness and suitability for multi-domain graph transfer.

\subsection{Analysis of Multi-domain Pre-training}

In Section ~\ref{sec:motivational_study}, we have demonstrated that incorporating more pretraining domains, particularly by mixing homogeneous and heterogeneous graphs, can substantially improve downstream performance on heterogeneous graphs. To further examine the generality and boundary of this observation, we extended this analysis to homogeneous graphs. The results are reported in Table ~\ref{tab:motivational_study_homo}.

From the results, we observed that increasing the number of pre-training graphs generally improved downstream performance on Cora and PubMed, indicating that pre-training domain diversity remains an important factor. Notably, when replacing Photo and Computer with other academic networks, i.e., ACM and DBLP, we observed an improvement in downstream performance on the Cora. This indicates that incorporating heterogeneous graphs into the pre-training process can improve performance on downstream homogeneous graph tasks. However, we observed no performance improvement on PubMed. This might be because the encoder, pre-trained on heterogeneous graphs, captured richer cross-type semantics that might not be fully useful for homogeneous tasks with a single node type. Importantly, this observation does not contradict our conclusion. We emphasize that homogeneous and heterogeneous graphs are not mutually exclusive in the pre-training stage, and that an appropriate combination of both can enhance downstream performance.

\begin{table}[!t]
\centering
\caption{Performance of 3-shot node classification under different pre-training graphs on Cora and PubMed.}
\label{tab:motivational_study_homo}
\resizebox{0.97\linewidth}{!}{
\begin{tabular}{clcc}
\toprule
\               			              & \textbf{Pre-training Graphs}  & \textbf{Micro-F1}     & \textbf{Macro-F1} \\ 
\midrule

\multirow{4}{*}{\textbf{Cora}}      & CiteSeer                       & 59.66$\pm$3.46 & 58.78$\pm$4.32 \\
                                    & CiteSeer PubMed                & 62.56$\pm$2.64 & 61.39$\pm$3.64 \\
                                    & CiteSeer PubMed Photo Computer & 63.59$\pm$3.54 & 62.10$\pm$4.19 \\
                                    & CiteSeer PubMed ACM DBLP       & 65.61$\pm$2.04 & 63.97$\pm$3.21 \\
\midrule
\multirow{4}{*}{\textbf{PubMed}}    & Cora                           & 58.06$\pm$5.57 & 57.53$\pm$5.86 \\
                                    & Cora CiteSeer                  & 58.97$\pm$7.24 & 57.88$\pm$7.81 \\
                                    & Cora CiteSeer Photo Computer   & 62.96$\pm$8.78 & 62.11$\pm$9.88 \\
                                    & Cora CiteSeer ACM DBLP         & 62.14$\pm$7.63 & 61.33$\pm$8.63 \\				 
\bottomrule
\end{tabular}
}
\end{table}

\begin{table}[th!]
\centering
\caption{Performance of 3-shot node classification under mixed graph-type pre-training settings.}
\label{tab:mix_type}
\resizebox{0.97\linewidth}{!}{
\begin{tabular}{cllcc}
\toprule
& \textbf{Pre-training Graphs} & \textbf{Methods} & \textbf{Micro-F1} & \textbf{Macro-F1} \\ 
\midrule
\multirow{3}{*}{\textbf{Cora}} & \multirow{3}{*}{CiteSeer PubMed ACM DBLP} & MDGPT & 49.22 & 48.31 \\
                               &                                           & MDGFM & 52.15 & 37.04 \\
                               &                                           & $\text{GPH}^{2}$ & 65.61 & 63.97 \\
\midrule
\multirow{2}{*}{\textbf{ACM}}  & \multirow{2}{*}{DBLP Cora CiteSeer}       & HGPrompt & 63.13 & 62.13 \\
                               &                                           & $\text{GPH}^{2}$  & 83.29 & 82.76 \\
\bottomrule
\end{tabular}
}
\end{table}

\subsection{Analysis of Mixed Graph-Type Pre-training}
To evaluate the performance under more challenging mixed homogeneous–heterogeneous multi-domain pre-training settings, we construct the following two cases:

\hangindent=1.0em
\hangafter=1
\subparagraph{1.} CiteSeer, PubMed, ACM, and DBLP are used as pre-training graphs, with Cora as the downstream task;
\hangindent=1.0em
\hangafter=1
\subparagraph{2.} DBLP, Cora, and CiteSeer are used as pre-training graphs, with ACM as the downstream task.
\hangindent=1.0em
\hangafter=1

Since no existing graph pre-training method explicitly supports mixtures of homogeneous and heterogeneous graphs, we select representative baselines and apply minimal modifications to adapt them to these settings. For Case 1, we compare with MDGPT~\cite{MDGPT} and MDGFM~\cite{MDGFM}, where heterogeneous graphs are converted into multiple independent homogeneous graphs based on meta-paths as input. For Case 2, we compare with HGPrompt~\cite{hgprompt}, where homogeneous graphs are directly treated as graph templates.

As shown in Table~\ref{tab:mix_type}, $\text{GPH}^{2}$ significantly outperforms the baselines in both cases. This is likely because MDGPT and MDGFM cannot fully exploit the semantic information in heterogeneous graphs during pre-training, while HGPrompt is unable to extract sufficiently informative knowledge from simple homogeneous graphs for downstream heterogeneous tasks. In contrast, the Unified Multi-View Graph Construction in $\text{GPH}^{2}$ not only performs multi-view augmentation for homogeneous graphs, but also transforms heterogeneous graphs into a unified representation form. The pre-training encoder further learns shared pre-training knowledge from these views. As a result, $\text{GPH}^{2}$ enables more effective knowledge transfer across graph types and domains.

\subsection{Ablation Study}
In this section, we conduct ablation studies to examine the contribution of the key components in $\text{GPH}^{2}$, including:

\begin{itemize}[leftmargin=*]
\item \textbf{w/o Expert}, where the domain-specific expert encoding is removed and a single encoder is used to train on all graphs;
\item \textbf{w/o Align}, where the expert-specific alignment is removed in the downstream stage, i.e., Eq. ~\ref{eq:alignment};
\item \textbf{w/o Attn}, where the task-oriented attention-based voting mechanism is replaced by simple uniform averaging, i.e., Eq. ~\ref{eq:attention_vectors} - Eq. ~\ref{eq:attention_fusion};
\item \textbf{w/o Ortho}, where the orthogonality constraint on the attention vectors is removed from the task-oriented fusion, i.e., Eq. ~\ref{eq:orthogonality}.
\end{itemize}

\begin{figure}[t]
  \centering
  \begin{minipage}[t]{0.48\columnwidth}
    \centering
    \includegraphics[width=\linewidth]{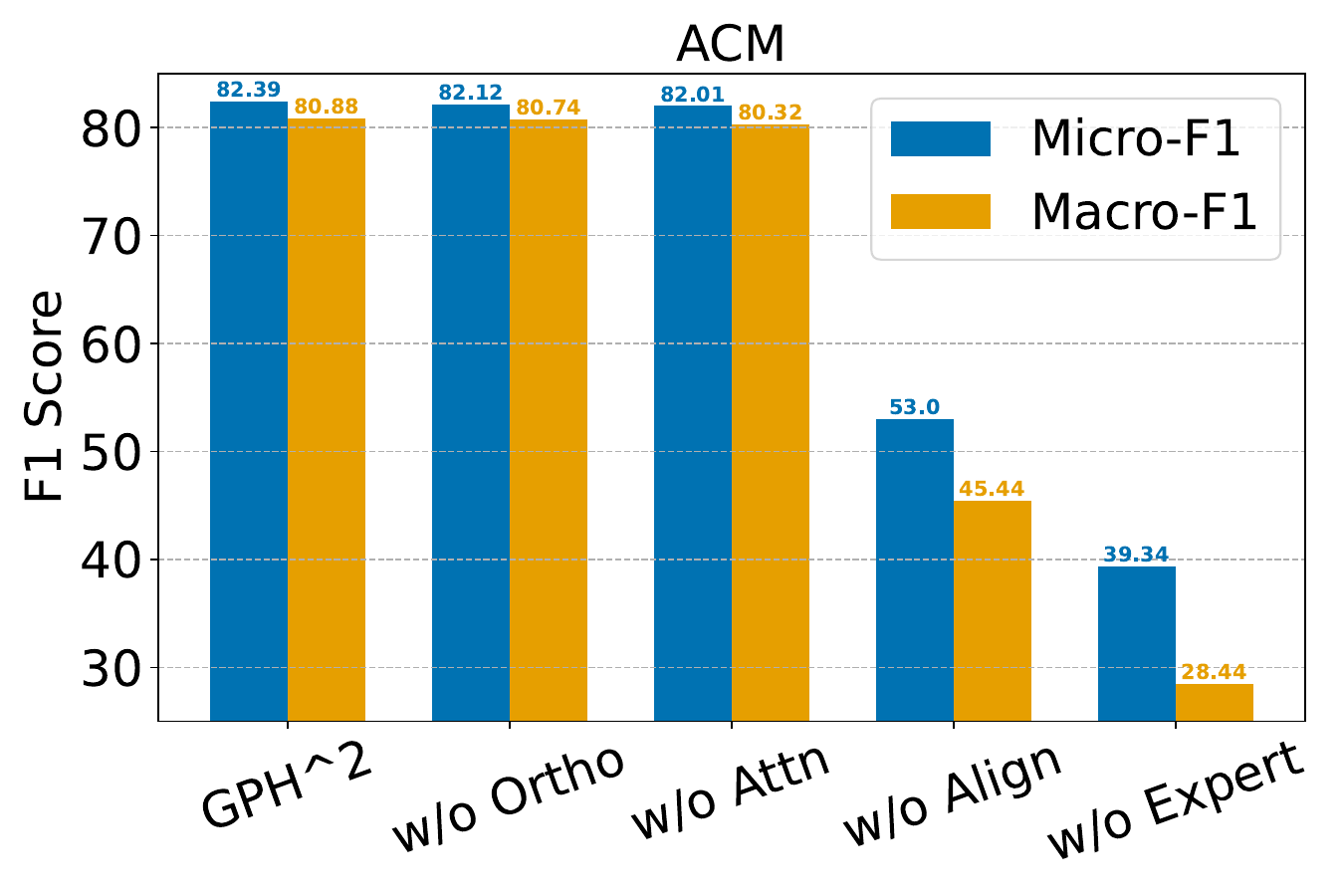}
  \end{minipage}
  \hfill
  \begin{minipage}[t]{0.48\columnwidth}
    \centering
    \includegraphics[width=\linewidth]{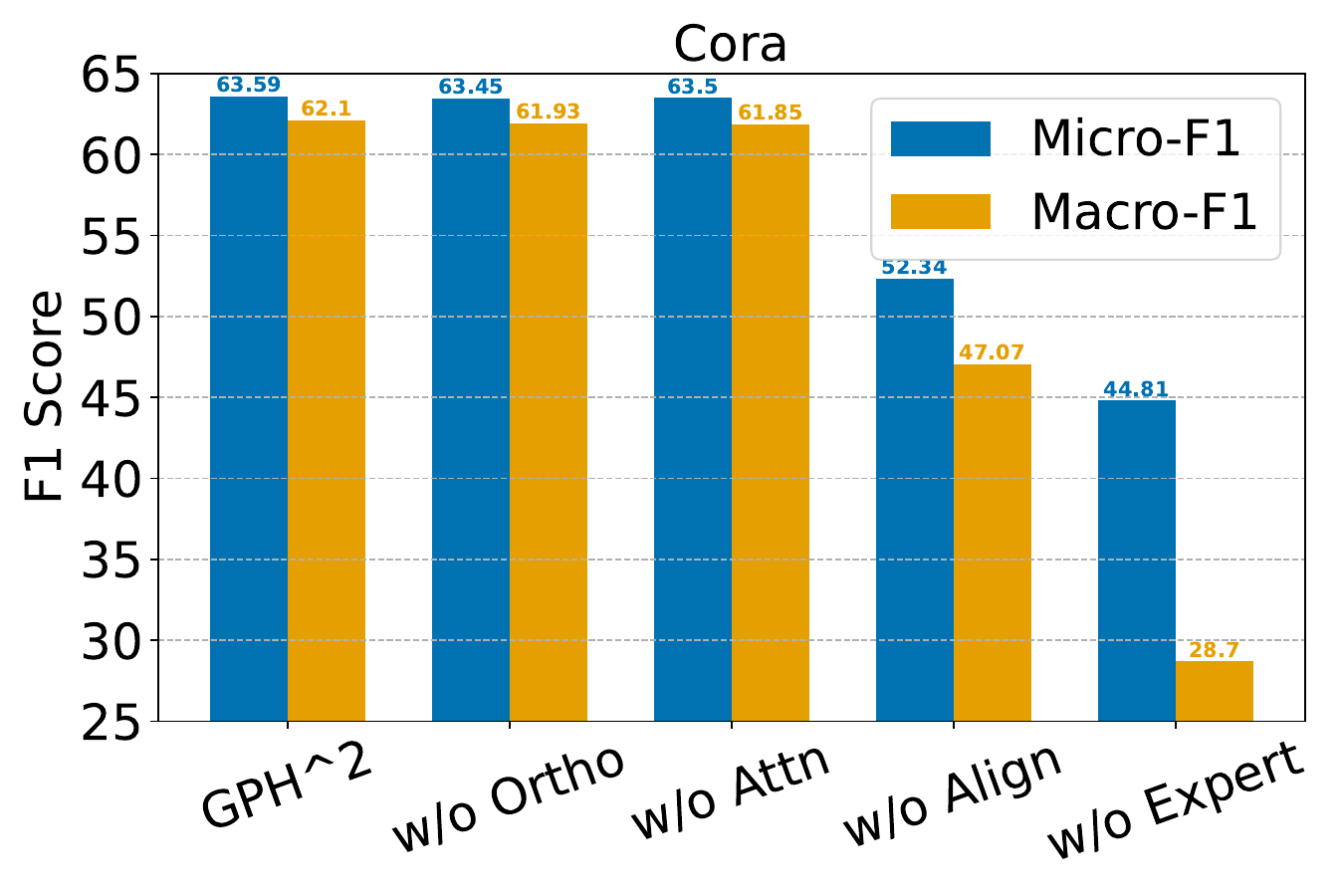}
  \end{minipage}
  \caption{Ablation study of on ACM and Cora.}
  \label{fig:ablation}
\end{figure}

All ablation experiments were conducted on the $3$-shot node classification tasks on ACM and DBLP. Based on the results in Figure ~\ref{fig:ablation}, we made the following observations. First, removing the domain-specific expert encoding (w/o Expert) resulted in a substantial performance drop on both ACM and Cora, indicating that domain-specific pretraining is critical for addressing cross-domain discrepancies. Second, removing expert-specific alignment (w/o Align) consistently degraded performance, suggesting that aligning expert representations into a shared space is necessary for effective downstream transfer, since $\text{GPH}^{2}$ performs no alignment operations at the upstream stage. Finally, removing either the task-oriented attention fusion or the orthogonality constraint resulted in a slight but consistent performance degradation. They helped stabilize expert integration by more effectively capturing experts' discriminative strengths, although their impact is less critical than that of domain-specific expert encoding and expert alignment.

\begin{figure}[t]
  \centering
  \includegraphics[width=\linewidth]{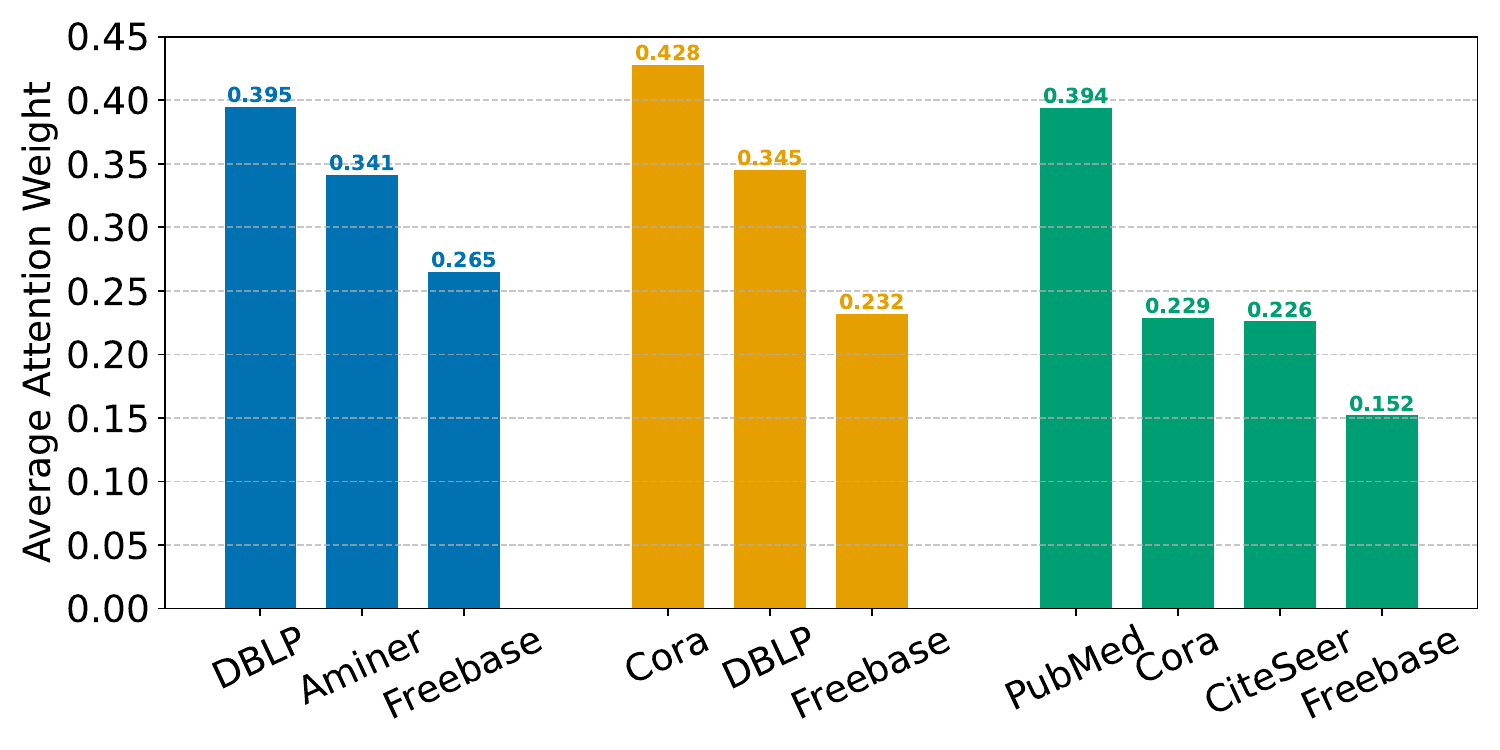}
  \caption{Attention scores of the Task-Oriented Expert Fusion module on ACM (50-shot node classification). The three colors represent three independent experiments.}
  \label{fig:attention}
\end{figure}

\subsection{Analysis of Task-Oriented Fusion}
To better understand how task-oriented expert fusion works in practice, we analyzed the learned attention weights of different experts on ACM, which reflect the contributions of pre-trained graphs. Since the attention weights are learned under label supervision, we conducted this analysis in a 50-shot setting. When available labels reached $50$ per class, the accuracy exceeded $90\%$, ensuring more reliable attention scores and reducing noise interference.

From Figure ~\ref{fig:attention}, we observed a consistent pattern: experts pre-trained on academic networks (e.g., DBLP, Aminer, Cora, CiteSeer, and PubMed) had substantially higher attention weights than experts pre-trained on Freebase (Freebase is a knowledge graph about movies). This trend held across all combinations of pre-training graphs, indicating that the attention mechanism can systematically select experts whose pre-training graphs are more relevant to ACM in terms of semantic and structural characteristics. Thereby validating the effectiveness of our task-oriented expert fusion design.

\subsection{Analysis of Pre-training Efficiency}
We further analyze the pre-training efficiency of $\text{GPH}^{2}$ in terms of convergence behavior and parameter size. We select DGI and DMGI as representative baselines since they share a similar objective with $\text{GPH}^{2}$, i.e., maximizing mutual information between local node representations and global graph summaries, while keeping the experimental settings consistent.

For homogeneous graph pre-training (CiteSeer, PubMed, Photo, and Computer as pre-training graphs, with Cora as the downstream task), DGI contains 33K parameters and requires 2,057 training epochs to converge. In comparison, $\text{GPH}^{2}$ contains 41K parameters and converges within 500 epochs for all source graphs (254, 498, 428, and 439 epochs, respectively). Although $\text{GPH}^{2}$ introduces a slight increase in parameters, it achieves substantially faster convergence.

For heterogeneous graph pre-training (DBLP, Aminer, and Freebase as pre-training graphs, with ACM as the downstream task), DMGI contains 1.86M parameters and still fails to converge within 3,000 epochs. In contrast, $\text{GPH}^{2}$ maintains only 41K parameters and converges within 800 epochs for all source graphs (340, 735, and 416 epochs, respectively).

These results demonstrate that the efficiency of $\text{GPH}^{2}$ is not hindered by the multi-expert design. In contrast, $\text{GPH}^{2}$ employs lightweight domain-independent expert, where each expert is trained separately, effectively avoiding interference across domains, thus accelerating convergence. Moreover, the expert encoders can be trained in parallel, further reducing the wall-clock time.

\section{Related Work}
\subsection{Homogeneous Graph Pre-training}
Homogeneous graph pre-training has been extensively studied~\cite{SSLsurvey1, SSLsurvey2}. Existing methods can be broadly categorized into generative self-supervised learning and contrastive self-supervised learning.

Generative-based methods aim to learn node or graph representations by reconstructing input features or structures from corrupted data. Early work, such as Graph Auto-Encoders (GAE)~\cite{GAE}, reconstructs adjacency matrices from latent embeddings. More recently, GraphMAE~\cite{graphmae}, GraphMAE2~\cite{graphmae2} and GPT-GNN~\cite{gpt-gnn} adopt a masked feature reconstruction objective, demonstrating strong scalability and transferability. These methods emphasize capturing global patterns through reconstruction-based objectives. Contrastive-based methods focus on learning representations by distinguishing positive pairs from negative pairs under different graph views. DGI~\cite{DGI}, GGD~\cite{GGD}, and MVGRL~\cite{mvgrl} adopt the principle of maximizing mutual information. They maximize global-local or view-level mutual information to preserve consistency in structure and semantics. Instead, GRACE~\cite{GRACE}, GraphCL~\cite{graphCL}, and GCC~\cite{gcc} adopt InfoNCE-style contrastive objectives, in which positive and negative samples are constructed via graph augmentation.

The above pre-training methods simply feed the learned embeddings into a classifier for predictions. More recently, inspired by prompt learning in large language models~\cite{nlppromptsurvey}, graph prompt-based pre-training and adaptation methods~\cite{gnnprompt1, gnnprompt2} have been proposed. These methods introduce tunable prompt parameters into graph data or a pre-trained encoder while keeping the encoder frozen during downstream training, yielding lightweight and effective adaptation. For example, GPF~\cite{gpf} and GPF-plus~\cite{gpf} inject global prompt vectors into node features, while SGL-PT~\cite{sgl-pt} and All-in-One~\cite{allinone} introduce virtual nodes or induced subgraphs as prompts. GraphPrompt~\cite{graphprompt} and MultiGPrompt~\cite{multigprompt} incorporate prompt tokens into pre-trained graph models to guide downstream adaptation. Beyond parameter tuning, most of these methods also reformulate downstream tasks to bridge the gap between pre-training and downstream task objectives. For example, GPPT~\cite{gppt} recasts node classification as a link-prediction task, whereas All-in-One~\cite{allinone} unifies node classification, link prediction, and graph classification into induced subgraph similarity estimation problems.

\subsection{Heterogeneous Graph Pre-training}
Heterogeneous graph pre-training has also been actively explored to capture rich semantic information induced by multiple node and edge types. Most existing approaches extend homogeneous graph pre-training by incorporating type- or schema-aware designs. For example, DMGI~\cite{DMGI} generalizes mutual information maximization to heterogeneous graphs by constructing meta-path-based views, while HeCo~\cite{HeCo} contrasts representations derived from network schema views and meta-path-based views to capture high-level cross-type semantics. More recently, HGMAE~\cite{HGMAE} extends masked autoencoding to heterogeneous settings by adopting type-specific masking and reconstruction strategies. The prompt-based method has also been explored for heterogeneous graphs. Methods such as HGPrompt~\cite{hgprompt} and HetGPT~\cite{hetgpt} introduce type-specific prompts to adapt heterogeneous graph encoders to downstream tasks.

Most heterogeneous graph pre-training methods rely on explicit schema, relation-specific encoders, or manually constructed meta-paths, which are tightly coupled to heterogeneous characteristics. Conversely, homogeneous graph pre-training methods typically use vanilla GNN encoders and cannot process heterogeneous graphs.

\subsection{Multi-domain Graph Pre-training}
Recent studies have explored multi- or cross-domain graph pre-training to learn transferable representations from diverse domains. SCOPE~\cite{SCOPE} amalgamates pre-training graphs by incorporating virtual nodes, enabling a single encoder to learn across multiple graph domains. MDGPT~\cite{MDGPT} proposes domain-aware prompt tokens to unify feature semantics during pre-training and further facilitates knowledge transfer by unifying and mixing prompts at the downstream stage. Similarly, SAMGPT~\cite{SAMGPT} designs structural prompts for structure alignment. MDGFM~\cite{MDGFM} further leverages graph structure learning to refine each pre-trained graph, integrating both feature and topological information during refinement.

However, existing multi-domain graph pre-training methods focus on homogeneous graphs, whereas heterogeneous graphs remain unexplored, let alone pre-trained across both homogeneous and heterogeneous graphs. It directly motivates our work, in which we propose $\text{GPH}^{2}$ to bridge this gap.

\section{Conclusion}
In this paper, we have revisited graph pre-training from a unified perspective, indicating the long-standing separation between homogeneous and heterogeneous graph pre-training. We empirically showed that combining homogeneous and heterogeneous graphs can learn more transferable representations. Motivated by this observation, we proposed $\text{GPH}^{2}$, a unified multi-domain graph pre-training framework that bridges homogeneous and heterogeneous graphs. $\text{GPH}^{2}$ is built upon three key designs: (i) a Unified Multi-View Graph Construction to reformulate diverse graph types into a common form, (ii) a domain-specific expert coding process to mitigate cross-domain distribution discrepancies, and (iii) a task-oriented expert fusion strategy to adaptively integrate domain-specific knowledge at the downstream stage. Together, these components enable stable and effective transfer across homogeneous and heterogeneous graphs, as confirmed by ablation analyses.

$\text{GPH}^{2}$ is a step toward truly unified graph foundation models that transcend graph-type boundaries and exhibit strong robustness to real-world domain heterogeneity and distribution shifts.

\begin{acks}
This work was supported by the National Natural Science Foundation of China (No.62422210, 62276187, 62532002, 92370111, 62272340), the Hong Kong RGC theme-based Strategic Target Grant Scheme (STG STG1/M-501/23-N), the Hong Kong Global STEM Professor Scheme, and the Hong Kong Jockey Club Charities Trust.
\end{acks}

\bibliographystyle{ACM-Reference-Format}
\bibliography{ref.bib}

\appendix

\section{Experiment Details}

\subsection{Datasets}
We evaluated $\text{GPH}^{2}$ on $9$ datasets covering multiple domains, including academic citation networks, product networks, and knowledge graphs. The statistics are summarized in Tables ~\ref{tab:datasets_ho} and ~\ref{tab:datasets_he}.

\subsection{Baselines}
We evaluate $\text{GPH}^{2}$ against 12 representative baselines across 4 categories: 
(1)~\textit{Homogeneous Self-Supervised Pre-training}: DGI~\cite{DGI}, GRACE~\cite{GRACE}, and GraphMAE~\cite{graphmae}; 
(2)~\textit{Homogeneous Graph Foundation Models}: GCOPE~\cite{SCOPE}, MDGPT~\cite{MDGPT}, and MDGFM~\cite{MDGFM}; 
(3)~\textit{Heterogeneous Self-Supervised Pre-training}: DMGI~\cite{DMGI}, HeCo~\cite{HeCo}, HGMAE~\cite{HGMAE}, and HERO~\cite{HERO}; 
(4)~\textit{Heterogeneous Prompt-based Fine-tuning}: HGPrompt~\cite{hgprompt} and HetGPT~\cite{hetgpt}. 

\begin{table}[th!]
\centering
\caption{The statistics of homogeneous datasets.}
\label{tab:datasets_ho}
\resizebox{0.7\linewidth}{!}{
\begin{tabular}{@{}lrrcc@{}}
\toprule
\# \textbf{Dataset}  & \# \textbf{Nodes}   & \# \textbf{Edges}    & \# \textbf{Classes} \\ 
\midrule
\textbf{Cora~\cite{Cora}}     & 2,708      & 10,556      & 7          \\
\textbf{Citeseer~\cite{citeseerPubmed}} & 3,327      & 9,104       & 6          \\
\textbf{PubMed~\cite{citeseerPubmed}}  & 19,717     & 88,648      & 3          \\
\textbf{Photo~\cite{photo}}    & 7,650      & 238,162     & 8          \\
\textbf{Computer~\cite{computer}} & 13,752     & 491,722     & 10         \\
\bottomrule
\end{tabular}
}
\end{table}

\begin{table}[th!]
\centering
\caption{The statistics of heterogeneous datasets. The underlined node type is the target type used for node classification.}
\label{tab:datasets_he}
\resizebox{1.0\linewidth}{!}{
\begin{tabular}{@{}cllcc@{}}
\toprule
					& \# \textbf{Nodes}             & \# \textbf{Edges}             & \textbf{Meta-Path}            & \# \textbf{Classes} \\ 
\midrule
\textbf{ACM~\cite{HeCo}}     	& \makecell*[l]{\underline{Paper (P)}: 4,019\\ Author (A): 7,167\\ Subject (S): 60} 					& \makecell*[l]{P-A:13,407\\ P-S:4,019} 				& \makecell*[c]{PAP\\ PSP} 				& 3	\\

\midrule
\textbf{DBLP~\cite{HeCo}}     	& \makecell*[l]{\underline{Author (A)}: 4,057\\ Paper (P):14,328\\ Conference (C):20\\ Term (T):7,723} 	& \makecell*[l]{P-A:19,645\\ P-C:14,328\\ P-T:85,810}   & \makecell*[c]{APA\\ APCPA\\ APTPA}	& 4	\\

\midrule
\textbf{Aminer~\cite{HeCo}}   	& \makecell*[l]{\underline{Paper (P)}: 6,564\\ Author (A):13,329\\ Reference (R):35,890} 			    & \makecell*[l]{P-A:18,007\\ P-R:58,831} 			    & \makecell*[c]{PAP\\ PRP}				& 4	\\

\midrule
\textbf{Freebase~\cite{HeCo}} 	& \makecell*[l]{\underline{Movie (M)}: 3;492\\ Actor (A):33,401\\ Direct (D):2,502\\ Writer (W):4,459} 	& \makecell*[l]{M-A:65,341\\ M-D:3,762\\ M-W:6,414} 	& \makecell*[c]{MAM\\ MDM\\ MWM}		& 3	\\ 
\bottomrule
\end{tabular}
}
\end{table}

\subsection{Implementation Details}
For all experiments, we fixed both the hidden and output dimensions to $128$. We also applied SVD~\cite{svd} to project all node feature dimensions onto $128$ dimensions. For all $m$-shot classification evaluation, we followed the ProG~\cite{ProG} benchmark settings. We conducted $10$ independent tasks by repeatedly sampling $m$ labeled nodes per class, and sampling $90\%$ of the nodes for testing. All experiments were conducted on a server equipped with an Intel Xeon(R) Platinum 8352V CPU and 90GB of RAM, accelerated by an NVIDIA RTX 4090 GPU with 24GB of VRAM.

For all baselines, we followed the original hyper-parameter settings in their code. Notably, some baselines (e.g., DMGI~\cite{DMGI}, HeCo~\cite{HeCo}) also rely on multi-view graph inputs. To adapt these methods to a multi-domain pre-training setting, where the number of graph views may vary across pre-training graphs, we modified their GNN encoders to share parameters across views, thereby enabling a fair comparison. For self-supervised pre-training baselines, we evaluated their learned representations through a linear classifier with Adam optimiser, while for others, we used their own downstream adaptation strategies.

During the pre-training stage, we adopted a $2$-layer SGC~\cite{sgc} as the GNN encoder of $\text{GPH}^{2}$, which serves as a lightweight and well-established backbone, allowing us to evaluate the effects of our key designs. 
For homogeneous graphs, we constructed $3$ views using edge dropping with a drop ratio of $0.3$. For heterogeneous graphs, the number of views is determined by the meta-paths listed in Table~\ref{tab:datasets_he}. $\text{GPH}^{2}$ is trained for up to $1,000$ epochs with early stopping (patience $= 50$), using a learning rate of $0.001$, dropout rate of $0.3$, and weight decay of $5\times10^{-4}$.
During the downstream stage, we froze the pre-trained encoder, and adjusted the depth of SGC for best performance. we fixed the number of epochs to $200$ and applied early stopping with a patience of $20$. The learning rate and dropout rate are selected from $\{0.05, 0.01, 0.005, 0.001, 0.0005\}$ and $\{0.0, 0.1, 0.2, 0.3, 0.4, 0.5, 0.6\}$, respectively. The best-performing hyper-parameters are reported in Table~\ref{tab:hyperparameters}.

\begin{table}[t!]
\centering
\caption{The best hyper-parameters.}
\label{tab:hyperparameters}
\resizebox{1.0\linewidth}{!}{
\begin{tabular}{cccccc}
\toprule
         & nb\_epochs & lr & wd & dropout & nb\_layers \\
\midrule
Cora     & 200  & 0.01      & $5\times10^{-4}$   & 0.2  &  5        \\
CiteSeer & 200  & 0.005     & $5\times10^{-4}$   & 0.2  &  4        \\
PubMed   & 200  & 0.01      & $5\times10^{-4}$   & 0.1  &  5        \\
Photo    & 200  & 0.01      & $5\times10^{-4}$   & 0.6  &  3        \\
Computer & 200  & 0.005     & $5\times10^{-4}$   & 0.2  &  2        \\
ACM      & 200  & 0.001     & $5\times10^{-4}$   & 0.2  &  2        \\
Aminer   & 200  & 0.05      & $5\times10^{-4}$   & 0.2  &  4        \\
DBLP     & 200  & 0.0005    & $5\times10^{-4}$   & 0.0  &  2        \\
Freebase & 200  & 0.0005    & $5\times10^{-4}$   & 0.2  &  4        \\
\bottomrule
\end{tabular}
}
\end{table}

\section{Additional Results}
\subsection{Hyper-parameter Analysis}

We investigate the sensitivity of two key hyper-parameters in homogeneous graph multi-view construction: the number of views and the drop-edge ratio. For heterogeneous graphs, the multi-view construction does not introduce additional hyper-parameters, as views are directly determined by predefined meta-paths.

\begin{figure}[th!]
  \centering
  \includegraphics[width=\linewidth]{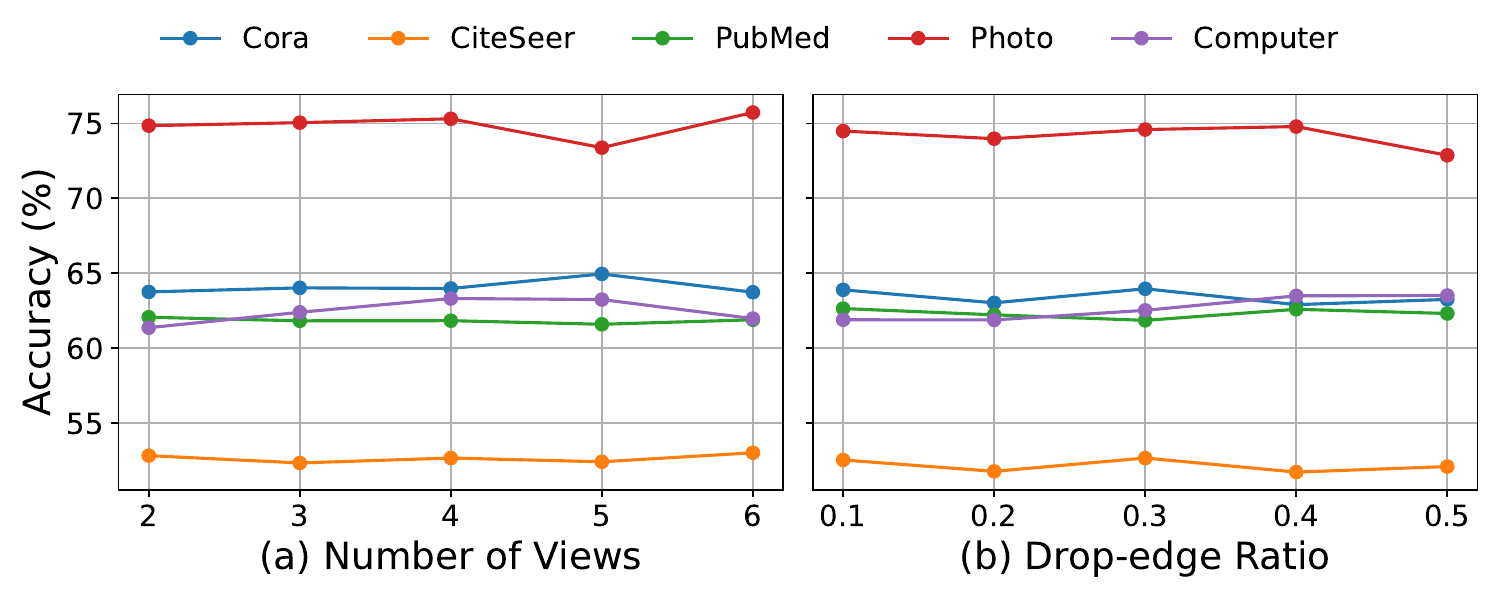}
  \caption{Analysis of the number of views and the drop-edge ratio in homogeneous graph multi-view construction.}
  \label{fig:param_analy}
\end{figure}

Figure ~\ref{fig:param_analy}(a) reports the results when fixing the drop-edge ratio to $0.3$ and varying the number of views from $2$ to $6$. Across all datasets, the performance remains relatively stable, with no clear trend as the number of views increases. In most cases, using $3$ or $4$ views achieves near-optimal performance.

Figure ~\ref{fig:param_analy}(b) presents the results when fixing the number of views to $3$ and varying the drop-edge ratio from $0.1$ to $0.5$. Overall, $\text{GPH}^{2}$ demonstrates strong robustness to different drop-edge ratios. While extremely small or large drop-edge ratios may slightly degrade performance on certain datasets, moderate perturbations (around $0.2-0.4$) consistently yield stable results.

Overall, these results indicate that $\text{GPH}^{2}$ is not sensitive to the choice of multi-view construction hyper-parameters. In practice, a small number of views combined with a moderate drop-edge ratio already provides reliable performance, supporting the robustness and practicality of the proposed multi-view construction strategy.


\subsection{Analysis of Task-Oriented Fusion}
\begin{figure}[th]
  \centering
  \includegraphics[width=\linewidth]{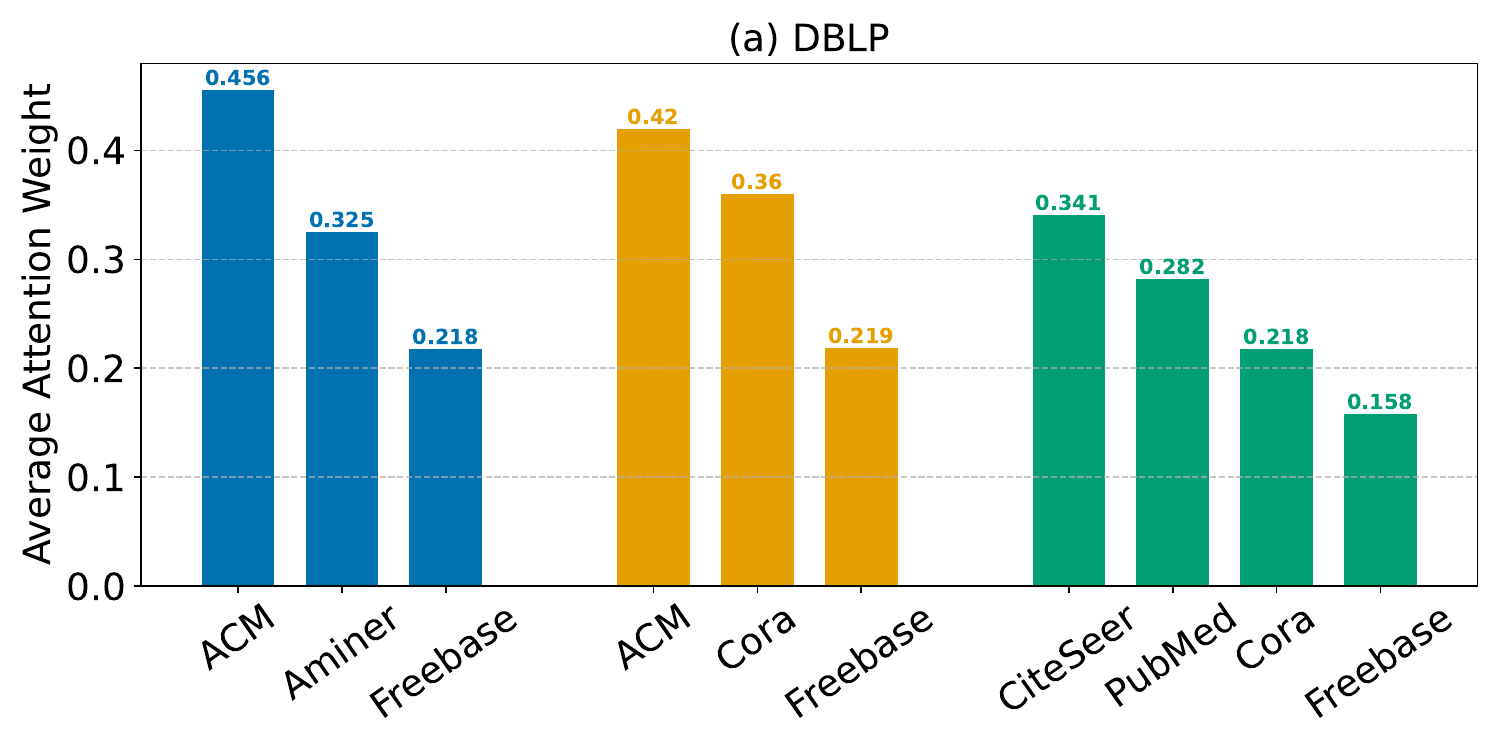}
  \includegraphics[width=\linewidth]{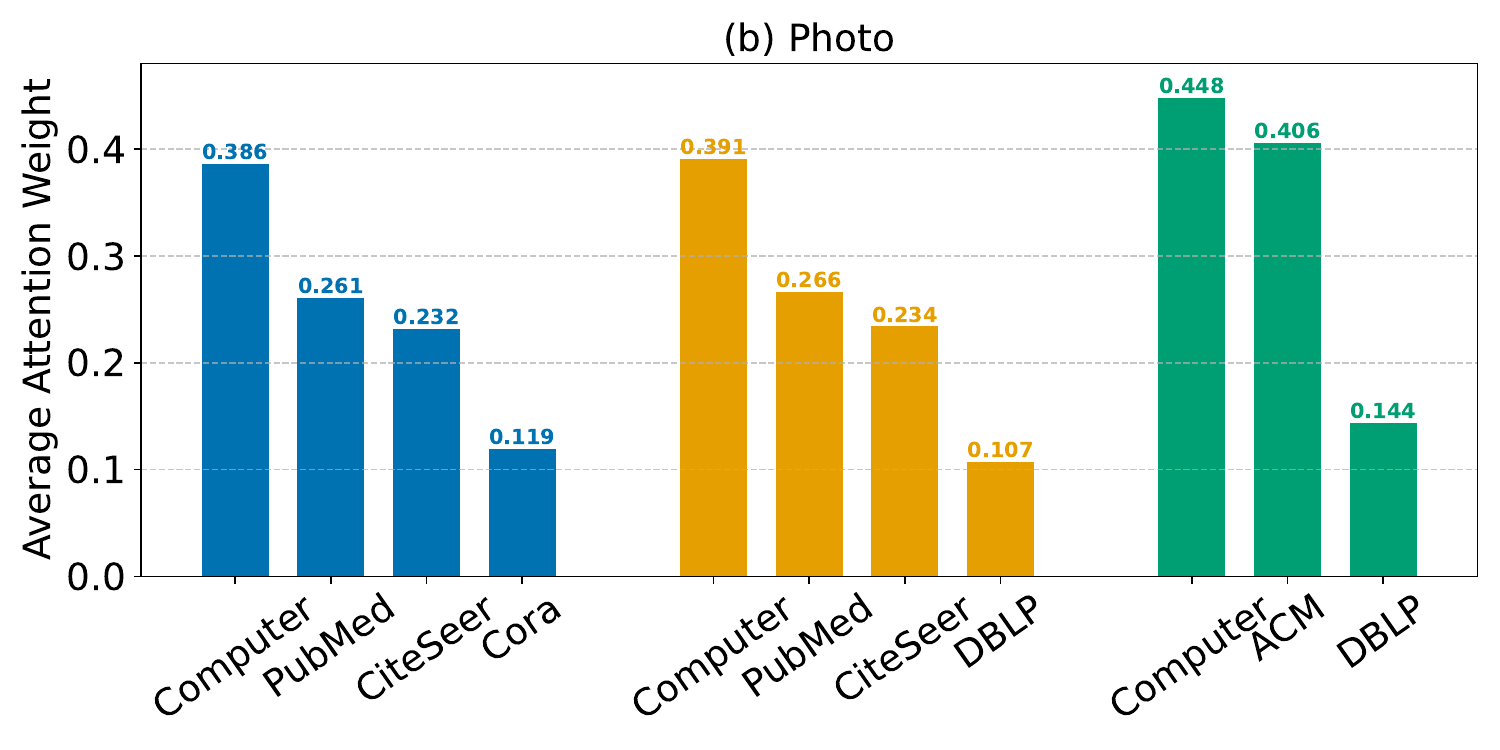}
  \caption{Attention scores of the Task-Oriented Expert Fusion module on Photo (50-shot node classification). The three colors represent three independent experiments.}
  \label{fig:attention_appendix}
\end{figure}

To further validate the generality of the task-oriented expert fusion mechanism, we extend the attention analysis to two additional downstream datasets, DBLP and Photo. We focus on these two datasets because, under the $50$-shot setting, their classification accuracy exceeds $90\%$, which ensures that the learned attention weights are more reliable and less affected by label noise. The results are shown in Figure ~\ref{fig:attention_appendix}.

We make two key observations:\textbf{(1) For DBLP}, expert pre-trained on ACM consistently received the highest attention weights, followed by other experts pre-trained on citation networks such as Aminer, Cora, CiteSeer, and PubMed. In contrast, expert pre-trained on Freebase was assigned substantially lower weights across different expert combinations. \textbf{(2) For Photo}, expert pre-trained on Computer (same as Photo as an e-commerce dataset) consistently dominated the attention weights across all expert groups. Notably, when comparing the blue and orange groups, replacing Cora with the higher-quality DBLP expert (which typically achieves higher classification accuracy than Cora) did not increase its attention weight. Instead, the attention assigned to DBLP decreased.

These results further indicate that the fusion mechanism selects experts whose pre-training graphs share stronger semantic and structural compatibility with the downstream graph, rather than favoring experts based on graph type or performance quality.

\end{document}